\documentclass{article}

      \usepackage[preprint]{tackling_climate_workshop_style}

\usepackage[utf8]{inputenc} 
\usepackage[T1]{fontenc}    
\usepackage{hyperref}       
\usepackage{url}            
\usepackage{booktabs}       
\usepackage{amsfonts}       
\usepackage{nicefrac}       
\usepackage{microtype}      
\usepackage{graphicx}

\title{An Industrial Workplace Alerting and Monitoring Platform to Prevent Workplace Injury and Accidents}

\author{%
  Sanjay Adhikesaven \\
  Foothill High School\\
  Pleasanton, CA 94568 \\
  \texttt{sanjay.adhikesaven1@gmail.com} \\
}

\begin{document}

\maketitle

\begin{abstract}
  Workplace accidents are a critical problem that causes many deaths, injuries, and financial losses. Climate change has a severe impact on industrial workers, partially caused by global warming. To reduce such casualties, it is important to proactively find unsafe environments where injuries could occur by detecting the use of personal protective equipment (PPE) and identifying unsafe activities. Thus, we propose an industrial workplace alerting and monitoring platform to detect PPE use and classify unsafe activity in group settings involving multiple humans and objects over a long period of time. Our proposed method is the first to analyze prolonged actions involving multiple people or objects. It benefits from combining pose estimation with PPE detection in one platform. Additionally, we propose the first open source annotated data set with video data from industrial workplaces annotated with the action classifications and detected PPE. The proposed system can be implemented within the surveillance cameras already present in industrial settings, making it a practical and effective solution. 

\end{abstract}

\section{Problem and Motivation}

Every year, almost two million people die from work-related accidents or diseases [1], and another 317 million suffer from work-related injuries [2]. The economic cost of these illnesses, injuries, and deaths is nearly three trillion dollars annually [3]. Beyond the direct toll on affected workers and medical expenses, workplace injuries lead to loss of productivity for all involved in the accident, property damage, and ultimately worsening morale among all workers [4]. Accounting for 39 percent of work related injuries, overexertion is the most common cause of these injuries [5]. Overexertion consists of lifting, lower, and other repetitive actions, which are all ergonomic issues. Another 31 percent of workplace injuries comes from contact with other objects [5], making the use of personal protective equipment (PPE) critical. 

Climate change has a strong impact on worker safety and health. Industrial workers are often most adversely affected by climate change and are exposed to physically demanding work for a longer period of time [6, 7]. Both outdoor and indoor factory workers face the impacts of increased heat [6, 7], which is often worsened by the lack of air conditioning and ventilation in many factories [8]. As global warming increases due to climate change, the impacts of extreme heat lead to many workplace injuries by lowering worker concentration and increasing fatigue [9]. For example, extreme heat has caused more than 20 thousands additional workplace injuries annually in California alone [9]. Overall, environmental heat causes over 170 thousand work-related injuries in the United States annually, making it the third highest cause of worker injuries [10]. It also contributes to as much as 2,000 deaths annually, possibly making heat exposure the leading cause of worker deaths [10]. Given the severe impact of climate change on workers, it is important to improve workplace safety and protect workers as a method of adapting to climate change. 

In this work, we propose an AI-powered industrial workplace alerting and monitoring platform in order to improve workplace safety. Our solution consists of two parts. First, we will detect personal protective equipment (PPE) compliance. Second, we will detect unsafe activity within the workplace using pose estimation. Our work will take a platform approach, allowing for easy implementation within the CCTV cameras already present within many industrial workplaces. Finally, we will release an open-source annotated dataset containing video data from industrial workplaces annotated with the action classifications and detected PPE. 

\section{Related Works}

Object detection is a technique commonly used in computer vision for a variety of problems. Significant prior work has been done to use object detection to detect the use of PPE [11-14]. [11, 12] use convolutional neural networks (CNNs) to detect the use of PPE (hard hat, vest, and shoes), while [13] can detect PPE (hard hat and vest) but also analyzes the safety level using the present PPE. If either a hard hat or vest is missing, then the output states the missing PPE. [14] considers PPE detection in the context of COVID-19 and detects PPE such as a face mask, face shield, and gloves. However, these prior methods have not been combined with activity detection methods to create a platform-based approach. Our proposed platform-based approach would allow industrial workplaces to have one system integrated within their cameras in order to monitor their workplace. 

Furthermore, pose estimation is widely used to study human action within a specific context. For analyzing unsafe activity within workplaces, previous work has focused on using pose estimation to monitor workplaces for ergonomic safety. [15] use deep learning to estimate the risk factor involved in lifting boxes, but cannot be generalized for more actions and does not account for the time taken per action. [16] use machine learning (random forest classifier) to classify activity in workplaces using sensors located on the foot but cannot determine when a set of actions are deemed to be unsafe. [17] use multiple machine learning algorithms to classify actions, but uses a time threshold to determine when a set of actions is unsafe. This does not account for human-object interactions and for the dynamic nature of time thresholds (lifting a heavy box for a period of time is different from lifting a light box for the same period of time). [18] can ergonomically analyze the actions of a single human but cannot differentiate between people and recognize human-object interactions. [19] determine the risk factor of a human’s pose but only looks at a single frame rather than a period of time. Our proposed method is the first to generalize unsafe activity detection to multiple people and/or objects over an extended period of time using pose estimation and action classification. 

Finally, there is no open-source dataset on industrial workplace actions, so we propose the first open-source annotated dataset containing images of industrial workers along with their classified action and detected PPE.  

\section{Proposed Approach}

We will annotate and release an open-source dataset based on real industrial workplace actions. We have collected raw data from video footage from sources such as Youtube, Shutterstock, Pixabay, and other web-based sources which contain videos of factories. After compiling the raw data, we have split the video into individual frames to find specific actions at a given time. We are in the process of annotating and labeling the dataset using Labelbox [20] with the correct action classification. The action will be annotated as slight bend, extended bend, random moves (small moves between actions), stand, or walk. Additionally, we will provide an annotation with respect to a given object, such as a box when such objects are present. In the case of multiple persons in a frame, we will assign the labels to the individual persons and denote their specific actions. Figure 1 shows two annotated frames from the proposed dataset. 

Our platform will combine PPE detection with unsafe activity detection to create a unified AI-based system that can be easily integrated within cameras in industrial workplaces. To perform PPE detection, we will use a pose estimation approach rather than an object detection approach, as used in [21]. This is motivated by the fact that, for the detection of PPE specifically, it is important to localize different body parts to know which protective equipment should be present. For example, a hard hat should be worn on the head, so the head must first be localized using pose estimation to determine whether a hard hat is present. A similar technique can be used for protective vest detection and shoe detection. 

In order to detect unsafe activity, we will use pose estimation to first classify individual poses. This can be done using OpenPose [22], which identifies a skeleton of 18 points at various key points that make up a human and will use those points to determine the action of the human through the distances and angles between the points. Using a multiple-target tracking (MTT) method as described in [23], each human in the frame will receive a label and their actions will be assigned to them. A report will be generated detailing each human’s action and the time taken per action, and this information can be used to determine whether a given set of actions is repetitive and can be deemed as unsafe. To detect human-object interactions, such as a human lifting a box or driving a forklift, we will use a human-object interaction detection method. We will experiment with multiple methods described in [24-26]. This will allow the platform to detect human-object interactions in addition to the specific action itself, which can be used to determine the risk level of a set of actions. After estimating the time taken and repetitive count for a set of actions per person, the data is then used to evaluate the postures using the Occupational Safety and Health Administration’s guidelines [27]. When a person’s actions are deemed as unsafe, an alert is sent to prevent potential future injuries.

\begin{figure}
  \centering
  \includegraphics[width = 1\columnwidth]{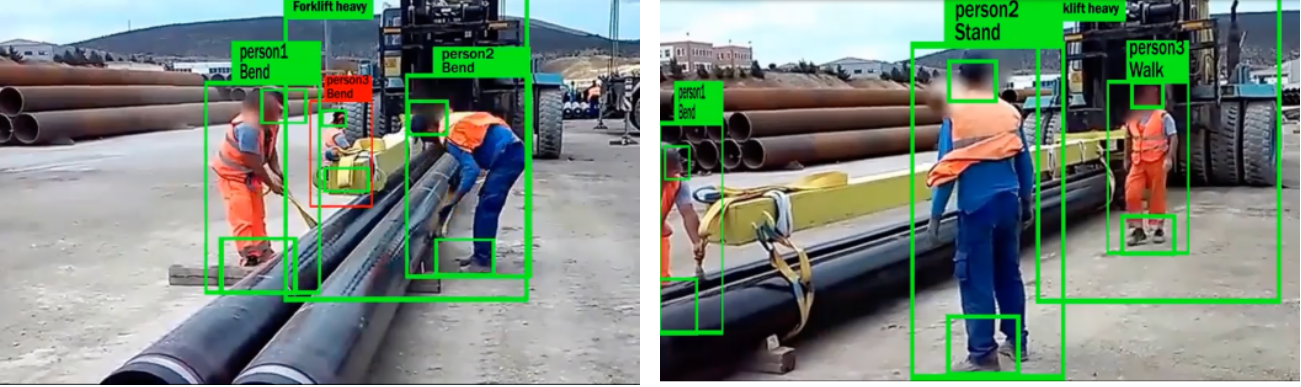}
  \caption{Example annotated images from the proposed dataset. Each individual is given a label, and bounding boxes are drawn around the detected PPE. The action classification is also shown under each person's label.}
\end{figure}

\section{Conclusion and Future Work}
\label{gen_inst}

In this paper, we proposed a platform to improve workplace safety. In addition, we are preparing an open source dataset to help train future models. Our method is the first to generalize the detection of unsafe activities to scenarios that contain multiple people and objects over a period of time. This can be done when multiple people and objects are within the same frame, and we will also account for tasks involving multiple people for the same task. Additionally, our method is the first to combine PPE detection and unsafe activity detection into a single platform that can be implemented in CCTV cameras to monitor industrial workplaces in real time.

Our future work will focus on deploying the platform within the CCTV cameras which are already present in factories today. Our platform does not require any specialized hardware other than a generic camera, making it a low-cost and practical solution to be implemented within industrial settings. Our proposal demonstrates the feasibility of using an AI-based platform to improve workplace safety in industrial settings, potentially helping the workers most severely affected by the impacts of climate change and saving many lives. 

\section*{References}

[1] World Health Organization. "WHO/ILO: Almost 2 Million People Die from Work-Related Causes Each Year." (2021).

[2] Woolford, Marta Helen, et al. "Missed opportunities to prevent workplace injuries and fatalities." New solutions: a journal of environmental and occupational health policy 27.1 (2017): 16-27.

[3] Safety+Health official magazine of National Safety Council of United State of America for Congress and Expo dated 06th September 2017 on Global cost of work-related injuries and deaths totals almost USD \$ 3 trillion. 

[4] DIRECT AND INDIRECT COSTS OF ACCIDENTS. Course 700 - Introduction to Safety Management. (n.d.). Retrieved September 4, 2022, from https://www.oshatrain.org/courses/pages/700costs.html 

[5] “Three Leading Causes of Workplace Injury and How to Prevent Them - South Shore Orthopedics.” South Shore Orthopedics, 15 Oct. 2021, southshoreorthopedics.com/three-leading-causes-of-workplace-injury-and-how-to-prevent-them.

[6] Kiefer, Max, et al. "Worker health and safety and climate change in the Americas: issues and research needs." Revista Panamericana de Salud Pública 40 (2016): 192-197.

[7] “Impact of Climate on Workers | NIOSH | CDC.” Impact of Climate on Workers | NIOSH | CDC, 6 Dec. 2016, www.cdc.gov/niosh/topics/climate/how.html.

[8] “Researcher Analyzes Effects of Climate Change on Productivity.” Harvard Gazette, 1 Nov. 2019, news.harvard.edu/gazette/story/2019/11/researcher-analyzes-effects-of-climate-change-on-productivity.

[9] “Work Injuries Tied to Heat Are Vastly Undercounted, Study Finds (Published 2021).” Work Injuries Tied to Heat Are Vastly Undercounted, Study Finds (Published 2021), 15 July 2021, www.nytimes.com/2021/07/15/climate/heat-injuries.html.

[10] “Boiling Point - Public Citizen.” Public Citizen, 28 June 2022, www.citizen.org/article/boiling-point.

[11] Nath, Nipun D., Amir H. Behzadan, and Stephanie G. Paal. "Deep learning for site safety: Real-time detection of personal protective equipment." Automation in Construction 112 (2020): 103085.

[12] Balakreshnan, Balamurugan, et al. "PPE compliance detection using artificial intelligence in learning factories." Procedia Manufacturing 45 (2020): 277-282.

[13] Delhi, Venkata Santosh Kumar, R. Sankarlal, and Albert Thomas. "Detection of personal protective equipment (PPE) compliance on construction site using computer vision based deep learning techniques." Frontiers in Built Environment 6 (2020): 136.

[14] Protik, Adban Akib, Amzad Hossain Rafi, and Shahnewaz Siddique. "Real-time Personal Protective Equipment (PPE) Detection Using YOLOv4 and TensorFlow." 2021 IEEE Region 10 Symposium (TENSYMP). IEEE, 2021.

[15] Liu, Ya, et al. "AI-based framework for risk estimation in workplace." Aggression and Violent Behavior (2021): 101616.

[16] Fridolfsson, Jonatan, et al. "Workplace activity classification from shoe-based movement sensors." BMC biomedical engineering 2.1 (2020): 1-8.

[17] Ray, Soumitry J., and Jochen Teizer. "Real-time construction worker posture analysis for ergonomics training." Advanced Engineering Informatics 26.2 (2012): 439-455.

[18] Martin, Chris C., et al. "A real-time ergonomic monitoring system using the Microsoft Kinect." 2012 IEEE Systems and Information Engineering Design Symposium. IEEE, 2012.

[19] Paudel, Prabesh \& Choi, Kyoung-Ho. (2020). A Deep-Learning Based Worker’s Pose Estimation. 10.1007/978-981-15-4818-5\_10. 

[20] "Labelbox | Software platform for building AI with unstructured data." N.p., n.d. Web. 13 Sep. 2022 https://labelbox.com

[21] Xiong, Ruoxin, and Pingbo Tang. “Pose Guided Anchoring for Detecting Proper Use of Personal Protective Equipment.” Automation in Construction, vol. 130, 2021, p. 103828., https://doi.org/10.1016/j.autcon.2021.103828. 

[22] Cao, Zhe, et al. "Realtime multi-person 2d pose estimation using part affinity fields."    Proceedings of the IEEE conference on computer vision and pattern recognition. 2017.

[23] Kamkar S, Ghezloo F, Moghaddam HA, Borji A, Lashgari R. Multiple-target tracking in human and machine vision. PLoS Comput Biol. 2020 Apr 9;16(4):e1007698. doi: 10.1371/journal.pcbi.1007698. PMID: 32271746; PMCID: PMC7144962.

[24] Ulutan, Oytun, A. S. M. Iftekhar, and Bangalore S. Manjunath. "Vsgnet: Spatial attention network for detecting human object interactions using graph convolutions." Proceedings of the IEEE/CVF conference on computer vision and pattern recognition. 2020.

[25] Gkioxari, Georgia, et al. "Detecting and recognizing human-object interactions." Proceedings of the IEEE conference on computer vision and pattern recognition. 2018.

[26] Kim, Bumsoo, et al. "Hotr: End-to-end human-object interaction detection with transformers." Proceedings of the IEEE/CVF Conference on Computer Vision and Pattern Recognition. 2021.

[27] “Department of Labor Logo United States Department of Labor.” Ergonomics - Identify Problems | Occupational Safety and Health Administration, https://www.osha.gov/ergonomics/identify-problems\#risk-factors.  

\end{document}